\documentclass[journal,twoside,web]{ieeecolor}

\usepackage{generic}
\usepackage{subcaption}

\usepackage{cite}
\usepackage{amsmath,amssymb,amsfonts}
\usepackage{algorithmic}
\usepackage{graphicx}
\usepackage{algorithm,algorithmic}
\usepackage{hyperref}
\hypersetup{hidelinks=true}
\usepackage{textcomp}
\def\BibTeX{{\rm B\kern-.05em{\sc i\kern-.025em b}\kern-.08em
    T\kern-.1667em\lower.7ex\hbox{E}\kern-.125emX}}
\begin{document}
\title{Information Entropy-Based Framework for Quantifying 
Tortuosity in Meibomian Gland Uneven Atrophy}
\author{Kesheng Wang, Xiaoyu Chen, Chunlei He, Fenfen Li, Xinxin Yu, Dexing Kong, Shoujun Huang, Qi Dai 
\thanks{This research was funded by the grants from Zhejiang Normal University 
(Grant Nos. YS304222929, YS304222977, PTK12923007, PTK12923012), and the National 
Natural Science Foundation of China (Grant Nos. 12090020 and 12090025), 
and the State Administration of traditional Chinese medicine Science and Technology 
Department-Zhejiang Provincial Administration of Traditional Chinese Medicine 
Co-construction Science and Technology Plan (Grant No. GZY-ZJ-KJ-23086). 
Corresponding authors: Shoujun Huang and Qi Dai.}
\thanks{Kesheng Wang, Chunlei He, Dexing Kong, Shoujun Huang and Qi Dai 
are with College of Mathematical Medicine, 
Zhejiang Normal University, Jinhua 321004,China, 
(e-mails: 2583510768@zjnu.edu.cn;chunlei@zjnu.edu.cn;dkong@zjnu.edu.cn; sjhuang@zjnu.edu.cn; dq@mail.eye.ac.cn). }
\thanks{Xiaoyu Chen, Fenfen Li, Xinxin Yu and Qi Dai are with National Clinical Research Center 
for Ocular Diseases, Eye Hospital, Wenzhou Medical University, Wenzhou 325027,China,
(e-mails: xiaoyuchenny@wmu.edu.cn;lifenfen@eye.ac.cn;13588775204@eye.ac.cn; dq@mail.eye.ac.cn). }}
\maketitle

\begin{abstract}
    In the medical image analysis field, precise quantification of 
    curve tortuosity plays a critical role in the auxiliary diagnosis and 
    pathological assessment of various diseases. In this study, we propose 
    a novel framework for tortuosity quantification and demonstrate its 
    effectiveness through the evaluation of meibomian gland atrophy uniformity, 
    serving as a representative application scenario.

    We introduce an information entropy-based tortuosity 
    quantification framework that integrates probability modeling with 
    entropy theory and incorporates domain transformation of curve data. 
    Unlike traditional methods such as curvature or arc-chord ratio, 
    this approach evaluates the tortuosity of a target curve by comparing 
    it to a designated reference curve. Consequently, it is more suitable 
    for tortuosity assessment tasks in medical data where biologically 
    plausible reference curves are available, providing a more robust 
    and objective evaluation metric without relying on idealized 
    straight-line comparisons.
    
    First, we conducted numerical simulation experiments to preliminarily 
    assess the stability and validity of the method. Subsequently, the 
    framework was applied to quantify the spatial uniformity of meibomian 
    gland atrophy and to analyze the difference in this uniformity between 
    \textit{Demodex}-negative and \textit{Demodex}-positive patient groups. The results 
    demonstrated a significant difference in tortuosity-based uniformity 
    between the two groups, with an area under the curve  of 0.8768, sensitivity of 0.75, 
    and specificity of 0.93. These findings highlight the clinical utility 
    of the proposed framework in curve tortuosity analysis and its potential as a 
    generalizable tool for quantitative morphological evaluation in medical diagnostics.

\end{abstract}

\begin{IEEEkeywords}
    \textit{Demodex} infestation, Information entropy, Meibomian gland atrophy
    , Morphological assessment, Tortuosity.
\end{IEEEkeywords}

\section{INTRODUCTION}
\label{sec:introduction}
\IEEEPARstart{T}{he} 
tortuosity of anatomical curves is crucial in clinical 
medical image analysis and has been widely applied in the auxiliary 
diagnosis and pathological evaluation of various 
diseases. For example, the tortuosity of retinal blood vessels 
in neonates serves as a crucial indicator for diagnosing retinopathy 
of prematurity, providing insights into the degree of 
abnormal neovascularization and disease progression\cite{chiang2021international}. 
Similarly, corneal nerve tortuosity plays an important role in the assessment 
of several neuropathic conditions, such as diabetic neuropathy and 
autoimmune disorders, as it reflects the status of nerve fiber 
damage and regeneration\cite{you2022application, williams2020artificial, 
wei2020deep, cruzat2017vivo}. 
In addition, for meibomian gland dysfunction (MGD), the tortuosity of the glandular boundary curves has 
been proposed as an indirect marker for evaluating the spatial uniformity of 
gland atrophy, thereby offering a quantitative basis for the early identification 
and intervention of related ocular surface diseases\cite{gao2005in, 
kheirkhah2007corneal, lee2010relationship, cheng2019correlation, 
randon2015confocal, alver2017clinical, zhang2018association, yan2020association}.

Previous studies have demonstrated a strong association between ocular 
\textit{Demodex} infestation and MGD, with the 
presence of mites shown to disrupt the normal morphology and function of the 
meibomian glands\cite{kheirkhah2007corneal, liang2014high, hao2022demodex}. 
The pathogenic mechanisms of \textit{Demodex} are multifactorial, 
including mechanical obstruction of glandular orifices, release of toxic metabolic 
byproducts, induction of inflammatory and immune responses, and synergistic 
effects with bacterial colonization—all of which contribute to ocular surface damage. 
Clinically, \textit{Demodex} infestation has been observed to induce characteristic focal 
or irregular atrophy of the meibomian glands\cite{gao2005in, kheirkhah2007corneal,
lee2010relationship, cheng2019correlation, randon2015confocal, alver2017clinical, 
zhang2018association, yan2020association}. However, despite these observations, 
the underlying mechanisms and structural manifestations of this nonuniform atrophy 
remain poorly understood, and there is currently no standardized method for its 
quantitative assessment. Therefore, developing objective and robust approaches to 
characterize the spatial distribution of glandular atrophy—particularly in relation 
to tortuosity-derived measures of structural uniformity—may offer valuable insights 
for the early detection of \textit{Demodex} infestation, facilitate more refined clinical 
evaluation, and support future investigations into its pathological mechanisms.

\section{RELATED WORK}
\label{sec:related work}
Currently, in vivo confocal microscopy (IVCM) is the clinical gold standard for diagnosing \textit{Demodex} 
blepharitis. However, this method 
requires a high level of patient cooperation and involves 
considerable time and cost, limiting its widespread use. The 
evaluation of meibomian gland atrophy uniformity associated 
with \textit{Demodex} blepharitis still primarily relies on subjective 
expert judgment, with a lack of effective and objective quantitative 
approaches\cite{saha2022automated, li2023unsupervised}, 
and related research remains limited\cite{liu2022uneven, lin2020novel}. 
Cross-disciplinary investigations have revealed 
that curve tortuosity 
assessment has various definitions and 
computational methods\cite{mao2020automated, smedby2008two, hart1999measurement}. 
Representative metrics include chord-to-arc ratio, curvature, 
least squares linear regression, and total variation, 
along with various extensions and improvements built upon these foundational 
techniques\cite{mao2020automated, hart1999measurement, sharafi2024automated, turior2013quantification, patasius2005evaluation, kashyap2022accuracy, dasilva2022analysis, wang2024improved, wang2017extracranial, badawi2022four, zebic2023coronary, krivosic2025assessment, huang2024computer}. 
These methods have been successfully applied 
in the analysis of medical images involving extracranial internal carotid 
arteries \cite{wang2017extracranial}, coronary arteries\cite{zebic2023coronary}, 
temporal arteries and 
veins\cite{huang2024computer}, 
as well as retinal vasculature\cite{sharafi2024automated, patasius2005evaluation, badawi2022four}.

However, these methods generally rely on comparing the target curve to 
an ideal straight line to quantify the degree of deviation, which presents 
inherent limitations in real-world clinical imaging scenarios. For instance, 
for meibomian glands, the boundary curves of healthy glands 
naturally exhibit a certain degree of physiological tortuosity. Therefore, using a 
straight line as the reference may fail to accurately reflect 
structural alterations. In contrast, employing a nonlinear reference is 
often more consistent with physiological reality. As a result, traditional 
approaches may be limited in their ability to distinguish pathological 
tortuosity from normal anatomical variation.

To address this, we propose a novel information entropy-based framework (IEBF) 
for evaluating curve tortuosity.
This method structurally compares the target 
curve under evaluation with a predefined standard curve and quantifies the 
degree of disorder in their difference space using information entropy, 
thereby assessing the tortuosity of the target curve. Furthermore, \textit{Demodex} 
blepharitis is used as a clinical validation scenario, where the tortuosity 
of the meibomian gland boundary curve serves as an indicator of meibomian 
gland atrophy uniformity. Additionally, the effectiveness of the proposed method in 
distinguishing \textit{Demodex} infection is evaluated. Experimental results demonstrate 
that the framework can effectively quantify the uniformity of meibomian gland 
atrophy and accurately differentiate between infected and non-infected cases. 
Compared to existing traditional methods, it shows significant advantages 
in clinical discriminative performance and exhibits promising potential 
for clinical application.

\section{MATERIALS}
\subsection{Data and Equipment}
This retrospective study was conducted at the Hangzhou campus of 
Wenzhou Medical University Eye Hospital from December 2023 to September 2024. 
It focused on patients who visited the hospital due to eye discomfort 
and underwent IVCM. Patients with negative 
results for \textit{Demodex} on this microscopy were included. Meanwhile, patients 
with \textit{Demodex} blepharitis, confirmed by positive mite infection on the same 
microscopy during the same period, were collected as controls.

According to the Chinese 
expert consensus on diagnosis and treatment of \textit{Demodex} blepharitis, 
the diagnostic criteria for \textit{Demodex} blepharitis include: (1) chronic 
or subacute bilateral ocular symptoms such as redness, itching, 
foreign body sensation, or recurrent, refractory chalazion; (2) 
abnormal eyelashes with sebaceous sleeve-like secretions at the 
eyelash roots (a diagnostic indicator), possibly accompanied by 
eyelid margin hyperemia and thickening; and (3) positive \textit{Demodex} examination. 
Patients' meibography images were captured using the 
Keratograph 5M (K5M; Oculus, Wetzlar, Germany). 
The images of the right upper eyelid were included for analysis.

All images were acquired at a resolution of 1360 $\times$ 1024 and saved
in bmp format. After excluding images with poor quality or insufficient 
clarity for analysis, 35 patients were included in the 
\textit{Demodex}-negative group and 70 in the \textit{Demodex}-positive group.

The study was conducted in accordance with the Declaration of Helsinki (as revised in 2013). 
This study was approved by the Institutional Review Board (IRB) of the Eye Hospital, 
Wenzhou Medical University (IRB approval No. H2023-045-K-42) and the requirement 
for individual consent for this retrospective analysis was waived due to the 
retrospective nature. 

The parameter settings of the meibomian gland segmentation model, 
as well as the hardware configuration used for its training and testing, 
followed those reported by Zhang et al.\cite{zhang2022meibomian}. The development of the 
curvature algorithm and related experiments were conducted on hardware 
equipped with a 13th Gen Intel(R) Core(TM) i7-13700KF processor operating 
at 3.40 GHz. The software environment consisted of Windows 11 as the operating 
system, PyCharm Community Edition 2024.3.1 (JetBrains, Prague, Czech Republic) 
as the development platform, and Python 3.12.8 (Python Software Foundation, Wilmington, USA) 
as the programming language.

\subsection{STATISTICAL ANALYSIS}
All statistical analyses in this study were performed in the Python 
environment (version 3.12.8). Key libraries used include numpy and 
pandas for data processing, scipy.stats for statistical testing, 
matplotlib and seaborn for visualization, and sklearn.metrics for 
performance evaluation and receiver operating characteristic (ROC) analysis.

We calculated and performed descriptive statistical analysis on 
the tortuosity data of the two groups. Descriptive statistics were presented as 
mean ±standard deviation (SD) or median (interquartilerange,IQR).
The normality of the data distribution was assessed using the Shapiro-Wilk test. Based on 
the results of the normality test, appropriate methods were 
selected for intergroup comparison: an independent samples t-test 
was used if the data were approximately normally distributed; otherwise, 
the nonparametric Mann-Whitney U test was applied.

The distribution characteristics of the tortuosity data from the two 
groups were visually compared using boxplots. Additionally, to evaluate 
the effectiveness of the proposed framework in discriminating \textit{Demodex} 
blepharitis, ROC curves were plotted. 
The overall classification performance was quantified by calculating the 
area under the curve (AUC), and the optimal classification threshold was 
determined using the Youden index to achieve the best balance between 
sensitivity and specificity. This analysis confirmed that the proposed 
entropy-based metric possesses strong discriminative ability for 
identifying \textit{Demodex} infection status and holds promising clinical application value.

\section{METHODS}
\label{sec:methods}

\subsection{Boundary Extraction and Refinement}
This study employed the meibomian gland segmentation model 
developed by Zhang et al.\cite{zhang2022meibomian} 
to segment meibomian glands from meibography images. 
Based on the segmentation results, an effective boundary 
extraction method was proposed. 

Panels (a) and (c) of Fig~\ref{fig1} show relatively uniform 
gland distribution in the \textit{Demodex}-negative group, whereas panels (b) and (d) 
represent the \textit{Demodex}-positive group, displaying characteristic uneven atrophy. 
Each image consists, from top to bottom, of the refined gland segmentation map, 
tarsal plate region map, and original image. The red curves indicate the
extracted and optimized boundaries of the meibomian glands. 
The detailed processing steps are as follows:
\begin{figure}[!t]
    \centerline{\includegraphics[width=\columnwidth]{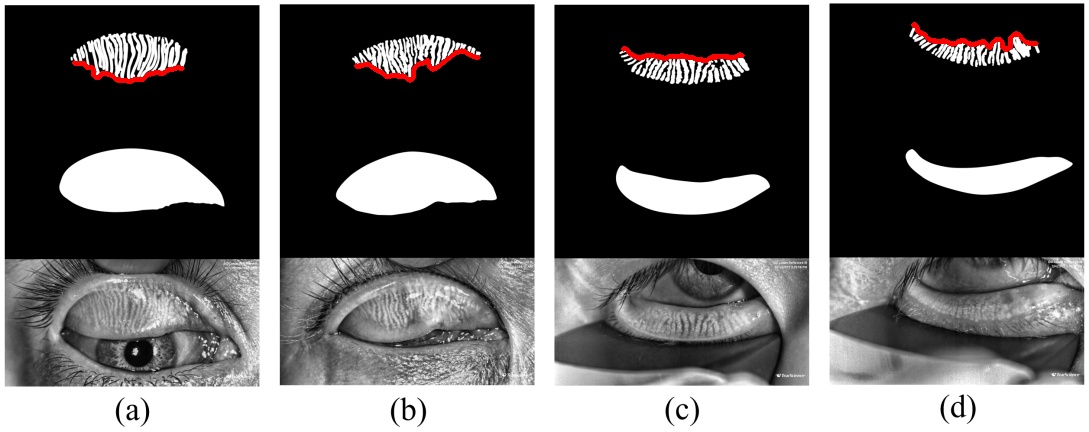}}
    \caption{ Meibomian gland segmentation and boundary extraction.
    (a, c): \textit{Demodex}-negative group; (b, d): \textit{Demodex}-positive group.
    Each panel shows (top to bottom): boundary map (red curve), tarsal region, and original image.}
    \label{fig1}
\end{figure}

First, the image is smoothed through Gaussian filtering using the 
following two-dimensional Gaussian kernel:
\begin{equation}
    G\left( i,j \right) =\frac{1}{2\pi \sigma ^2}\cdot e^{-\frac{i^2+j^2}{2\sigma ^2}}.
    \label{1}
\end{equation}

The image is convolved with the Gaussian kernel to obtain the blurred image
\begin{equation}
    I'\left( x,y \right) =\sum_{i=-k}^k{\sum_{j=-k}^k{G\left( i,j \right)}}\cdot I\left( x+i,y+j \right),
\end{equation}
where $i$ and $j$ represent the coordinate offsets relative to the center of the Gaussian kernel; 
$\sigma$ is the SD of the Gaussian kernel, $G(i,j)$ denotes the weight of 
the Gaussian kernel at position $(i,j)$; $I(i,j)$ and $I'(i,j)$ represent the grayscale 
values of the input and output (blurred) images at the corresponding positions, 
respectively. $k$ is the radius of the Gaussian kernel. In this study, the parameters were 
set as $\alpha = 0$ and $k=51$.These parameters were determined 
based on empirical results that yielded optimal performance.

The boundary information of the meibomian gland region is initially extracted from 
the output image. Subsequently, an active contour model is employed to further optimize 
the curve's position, where the energy function is defined by

\begin{equation}
    E=\int{\left( E_{internal}\left( s \right) +E_{external}\left( s \right) \right) \ ds},
\end{equation}
in which
\begin{equation}
    E_{internal}\left( s \right) =\alpha \lVert \frac{dv\left( s \right)}{ds} \rVert ^2+\beta \lVert \frac{d^2v\left( s \right)}{ds^2} \rVert ^2,
\end{equation}
\begin{equation}
    E_{external}\left( s \right) =-\left| \nabla I\left( v\left( s \right) \right) \right|.
\end{equation}

Note that $E_{internal}$ represents the smoothness and continuity of the curve, 
typically related to its bending degree; $E_{external}$ reflects 
the relationship between the curve and the image edges, 
encouraging the curve to move toward the true boundary of the 
meibomian gland region. To minimize the energy 
function, we need to compute the gradient of the total energy 
with respect to each point on the curve. 
The gradient indicates the direction and rate of change of the 
energy function at that point, guiding the 
movement of the curve. For each point $X_{i}=(s_{i},v(s_{i}))$, 
the gradient can be expressed as
\begin{equation}
    \nabla E\left( X_i \right) =\nabla E_{internal}\left( X_i \right) +\nabla E_{external}\left( X_i \right). 
\end{equation}

Each point on the curve is updated using the gradient descent method. 
By moving in the direction opposite to the gradient, the energy gradually decreases, 
and the curve evolves toward the position of minimum energy. The updated formula is given by
\begin{equation}
    X_{i,j+1}=X_{i,j}-\mu \nabla E\left( X_i \right), 
\end{equation}
where $X_{i}$ denotes the coordinate of the $i$-th point on the curve at the $t$ iteration, 
$\mu$ is the step size controlling the movement speed of the curve points during each iteration, 
and $\nabla E(X_{i})$ is the gradient of the energy function at point $X_{i}$. 
Through this updated formula, the curve gradually evolves toward the minimum energy state until 
convergence is reached, yielding the optimized boundary curve.
In this study, the parameters were set as $\alpha = 0.1$, $\beta = 1.0$ and $\mu = 0.1$.
These parameters were also determined based on empirical results that yielded optimal performance.

In this study, the extremal points of the optimized boundary curve are used 
to truncate the curve, 
retaining only the segment between the two extreme points. This process 
effectively adjusts the extracted
curve to better conform to the true boundary of the meibomian gland region 
while ensuring the smoothness and continuity of the curve. This approach significantly improves the 
accuracy of boundary extraction.

\subsection{Calculation of the Curve Tortuosity }
This study proposes that the evaluation of curve tortuosity requires 
a reference standard curve as a baseline. 
The standard curve is defined as an ideally unbent curve under a specific task, 
whereas the curve to be evaluated 
is referred to as the target curve. To quantify the difference between the 
target and baseline curves, 
we present an IEBF for tortuosity calculation.

Information entropy is a metric used to assess the uncertainty or disorder within 
a system. For a set of discrete 
data points, higher entropy indicates more greater variability and uncertainty among 
the points, whereas lower entropy suggests more uniformity. 
Mathematically, information entropy is typically defined by
\begin{equation}
    H\left( X \right) =-\sum_{i=1}^n{p\left( x_i \right) \cdot \log \left( p\left( x_i \right) \right)},
\end{equation}
where $H(X)$ denotes the information entropy of random variable $X$, 
$x_i$ represents the $i$-th possible value of $X$, $p(x_i)$
denotes the probability that $X$ takes the value $x_i$, and $n$ is the 
total number of possible values that $X$ can take.

When calculating the curve tortuosity, we evaluate the 
characteristics of distribution of point-wise distance 
differences between the target and baseline curves, 
and measure the tortuosity by quantifying the magnitude 
of disorder of these differences. The specific procedure is as follows.

\subsubsection{Construct the Point-to-Point Distance Difference Vector}

\begin{figure}[!t]
    \centerline{\includegraphics[width=\columnwidth]{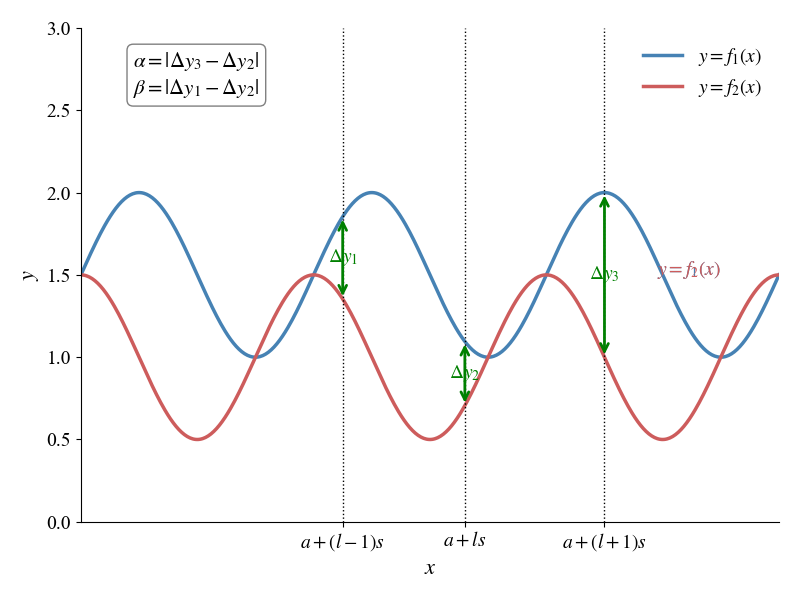}}
    \caption{Schematic diagram of the construction of the point-to-point distance difference vector.}
    \label{fig2}
    \end{figure}

Assume there are two discrete curves located on the same plane. The first 
curve is the standard curve, 
composed of $n$ coordinate points, and it is denoted by $y=f_{1}(x)$. 
The second curve is the target curve, 
also composed of $n$ coordinate points, and is denoted as $y=f_{2}(x)$. 
The domain for both curves is assumed to be the same 
(i.e., $[a,b]$). Let $s$ be a constant representing the sampling interval. Introduce the following notations: 
\begin{equation}
    \delta_k = f_1(a + ks) - f_2(a + ks),
\end{equation}
\begin{equation}
    \alpha = \lVert \delta_{l+1} \rVert - \lVert \delta_l \rVert,
\end{equation}
\begin{equation}
    \beta = \lVert \delta_{l-1} \rVert - \lVert \delta_l \rVert,
\end{equation}
where $l\in Z^+$, $l\le \left[ \frac{b}{s} \right] =q$. The distance difference vector is then
\begin{equation}
    d_l=\frac{1}{2}\lVert \alpha \rVert +\frac{1}{2}\lVert \beta \rVert.
\end{equation}
We collect all the distance differences in the form of a vector
\begin{equation}
    D=\left( d_1,\ d_2,\ \cdots ,\ d_q \right). 
\end{equation}
The schematic diagram of this process is shown in Fig~\ref{fig2}.

\subsubsection{Construction of the Probability Vector}
The standard Gaussian probability density function is given by
\begin{equation}
    f\left( d \right) =\frac{1}{\sqrt{2\pi \sigma ^2}}e^{-\frac{\left( d-\mu \right) ^2}{2\sigma ^2}},
\end{equation}
where we take $\mu=0$, $\sigma=1$. The corresponding probability function is expressed as
\begin{equation}
    g\left( x \right) =\int_{-\infty}^x{f\left( t \right) \ dt}.
\end{equation}
Using the probability function, the distance vector can be transformed into a probability vector
\begin{equation}
    P=\left( g\left( d_1 \right) ,\ g\left( d_2 \right) ,\ \cdots ,\ g\left( d_q \right) \right). 
\end{equation}

\subsubsection{Construction of the Tortuosity Mapping}
Based on the above discussion, we introduce the concept of total tortuosity, 
which is computed using the following formula 
\begin{equation}
    \label{eq:17}
    e\left( P \right) =\left( -\frac{1}{q}\sum_{p=1}^q{\left( 1-2g\left( d_p \right) \right) \cdot \log \left( 2g\left( d_p \right) \right)} \right) ^{\frac{1}{2}}.
\end{equation}
This expression can be regarded as an entropy-based approximation 
function incorporating both distance and probability. Here, $e(P)$ 
denotes the entropy value, and $g(d_p)$ is a function applied to 
the distance difference $d_p$. Because the range of value $g(d_p)$ 
is $(0, \frac{1}{2})$, it is scaled by a factor of 2 to extend the range to $(0, 1)$. 
As $d_p$ increases, $g(dp)$ decreases; when $d_p=0$, 
we have $g(dp)=\frac{1}{2}$. Therefore, the term $plogp$ in the original information entropy
is reformulated as $(1-p)log p$, leading to Eq. \eqref{eq:17}

\subsection{Tortuosity Correction}
Generally, the real acquired data contains various types of noise, 
which could affect the tortuosity values. 
Therefore, the original data is 
transformed into frequency spaces by using Fourier 
transform to extract either low-frequency or 
high-frequency components. The extracted components are 
then transformed back to the spatial domain via inverse 
Fourier transform, and tortuosity calculation can be performed. 
This process allows to focus specifically on either the 
low- or high-frequency information of the curve.

\subsubsection{Fourier Transform}

Given a one-dimensional time-domain signal, its Fourier transform is defined as
\begin{equation}
    F\left( \omega \right) =\frac{1}{2\pi}\int_{-\infty}^{+\infty}{f\left( x \right) e^{-i\omega x}dx},
\end{equation}
where $f(x)$ is the time-domain signal, representing a function over time (or space). 
$F(\omega)$ is the frequency-domain signal, 
describing the distribution of the signal over different frequencies. 
$\omega$ is the angular frequency, 
which is related to frequency $\xi$ by $\omega = 2\pi\xi$, 
with units of radians per second. $i$ is the imaginary unit.

\subsubsection{Frequency Band Extraction}
Low-frequency components ($F_{\text{low}}(\omega)$) typically represent the overall 
trend or slowly varying part of the signal, whereas high-frequency 
components ($F_{\text{high}}(\omega)$) contain rapidly changing details or noise. 
By defining a low-frequency cutoff point ($\omega_{\text{L}}$), 
the low-frequency portion of the signal can be extracted; similarly, 
a high-frequency cutoff point ($\omega_{\text{H}}$) 
is used to isolate the high-frequency part. That is
\begin{equation}
    F_{low}\left( \omega \right) =\left\{ \begin{array}{c}
        F\left( \omega \right) ,\ if\left| \omega \right|\le \omega _L\\
        0,\ if\left| \omega \right|>\omega _L\\
\end{array}
    \right.,
\end{equation}
\begin{equation}
    F_{high}\left( \omega \right) =\left\{ \begin{array}{c}
        F\left( \omega \right) ,\ if\left| \omega \right|\ge \omega _H\\
        0,\ if\left| \omega \right|<\omega _H\\
\end{array} 
    \right.. 
\end{equation}

\subsubsection{Inverse Fourier Transform}
The extracted frequency bands (low- or high-frequency components) 
are transformed back to the 
original domain through the inverse Fourier transform. This allows us to 
observe the contribution of each 
frequency band within the original signal. Given a frequency-domain signal, 
the inverse Fourier transform 
converts it back to the time-domain signal. 
The formula for the inverse Fourier transform is
\begin{equation}
    f\left( x \right) =\frac{1}{2\pi}\int_{-\infty}^{+\infty}{F\left( \omega \right) e^{i\omega x}d\omega}.
\end{equation}

By integrating the analysis of both low- and 
high-frequency information, a comprehensive evaluation 
of tortuosity at global and local scales can be achieved. 
This analytical approach not only enhances the 
accuracy of tortuosity calculation, but also better captures 
the complex characteristics of the data, 
thereby providing more reliable support for research and applications in related fields.

\section{EXPERIMENTS AND RESULTS ANALYSIS}
\subsection{Numerical Experiment on Tortuosity Quantification}

Owing to the difficulty of obtaining clinical data and the complexity 
and time-consuming nature of the curve extraction process, this 
study first conducts numerical experiments to preliminarily 
validate the effectiveness and stability of the proposed method. 
After validation, the method is further applied to real 
clinical data for additional verification.

\begin{figure}[!t]
\centerline{\includegraphics[width=\columnwidth]{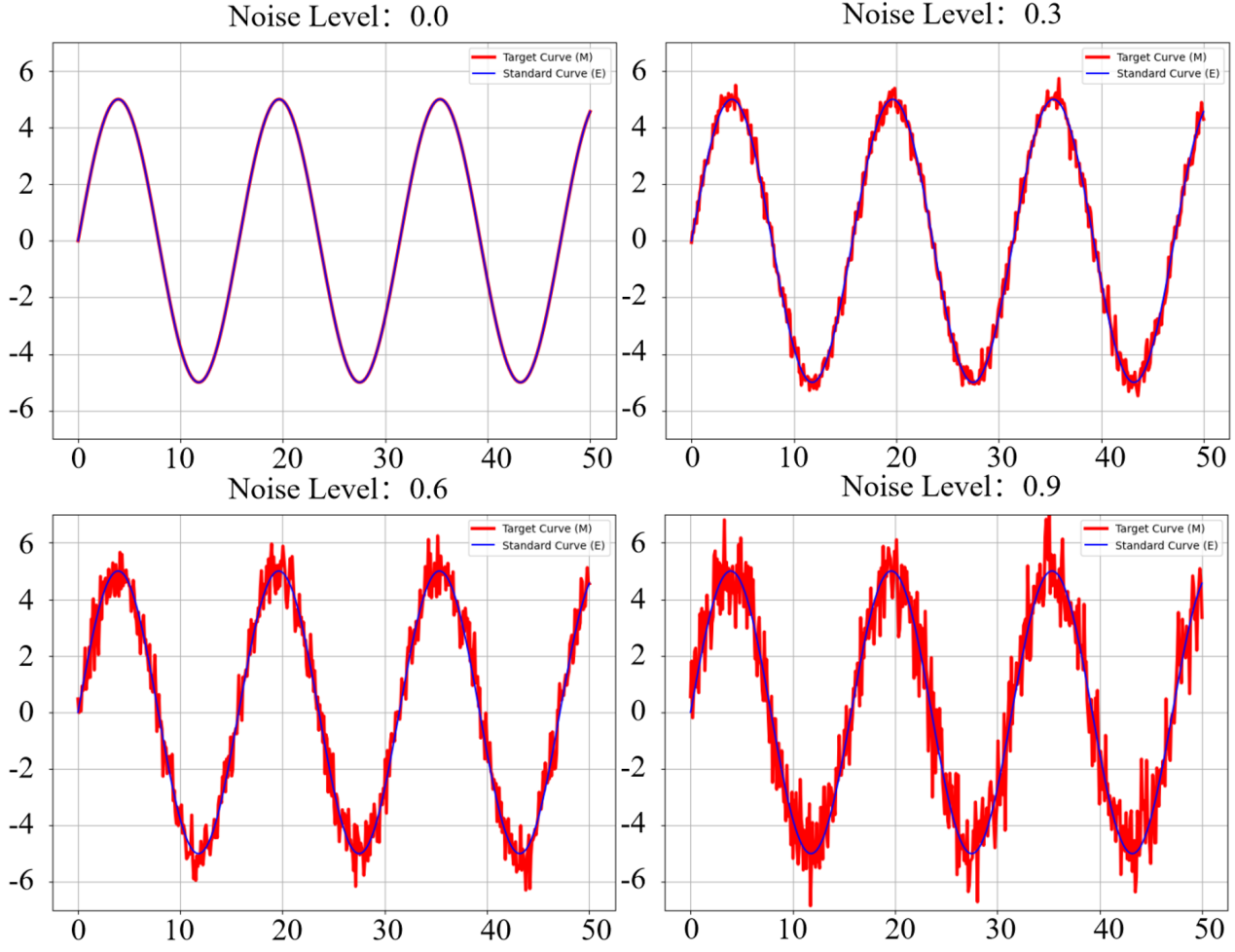}}
\caption{Comparison of target curves and standard 
curve under different levels of gaussian noise.}
\label{fig3}
\end{figure}

\begin{figure}[!t]
    \centering
    \begin{subfigure}[b]{\columnwidth}
        \centering
        \includegraphics[width=\columnwidth]{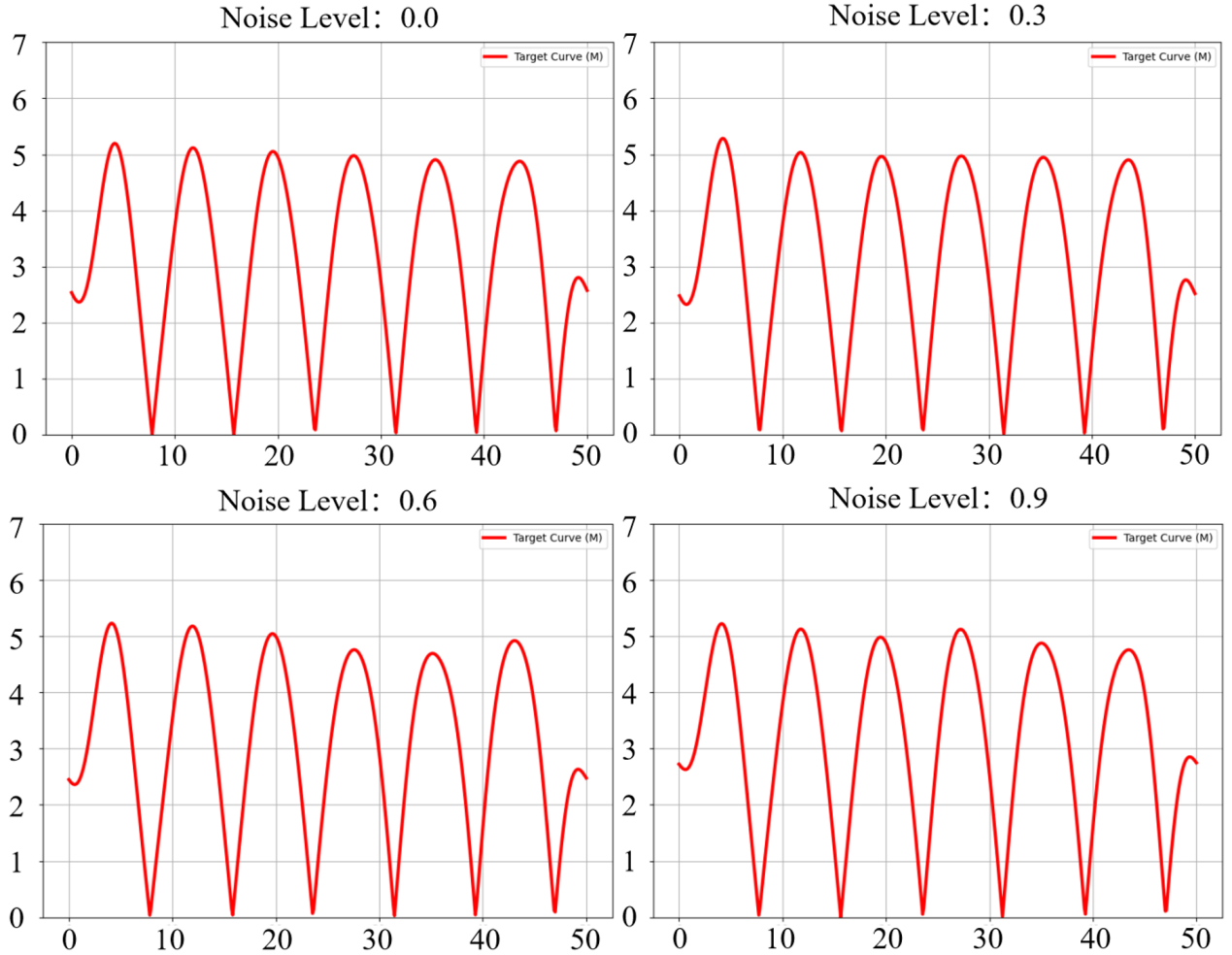}
        \caption{Low-frequency information curve}
        \label{fig4.1}
    \end{subfigure}

    \vspace{0.5em} % 可调整两个子图间距

    \begin{subfigure}[b]{\columnwidth}
        \centering
        \includegraphics[width=\columnwidth]{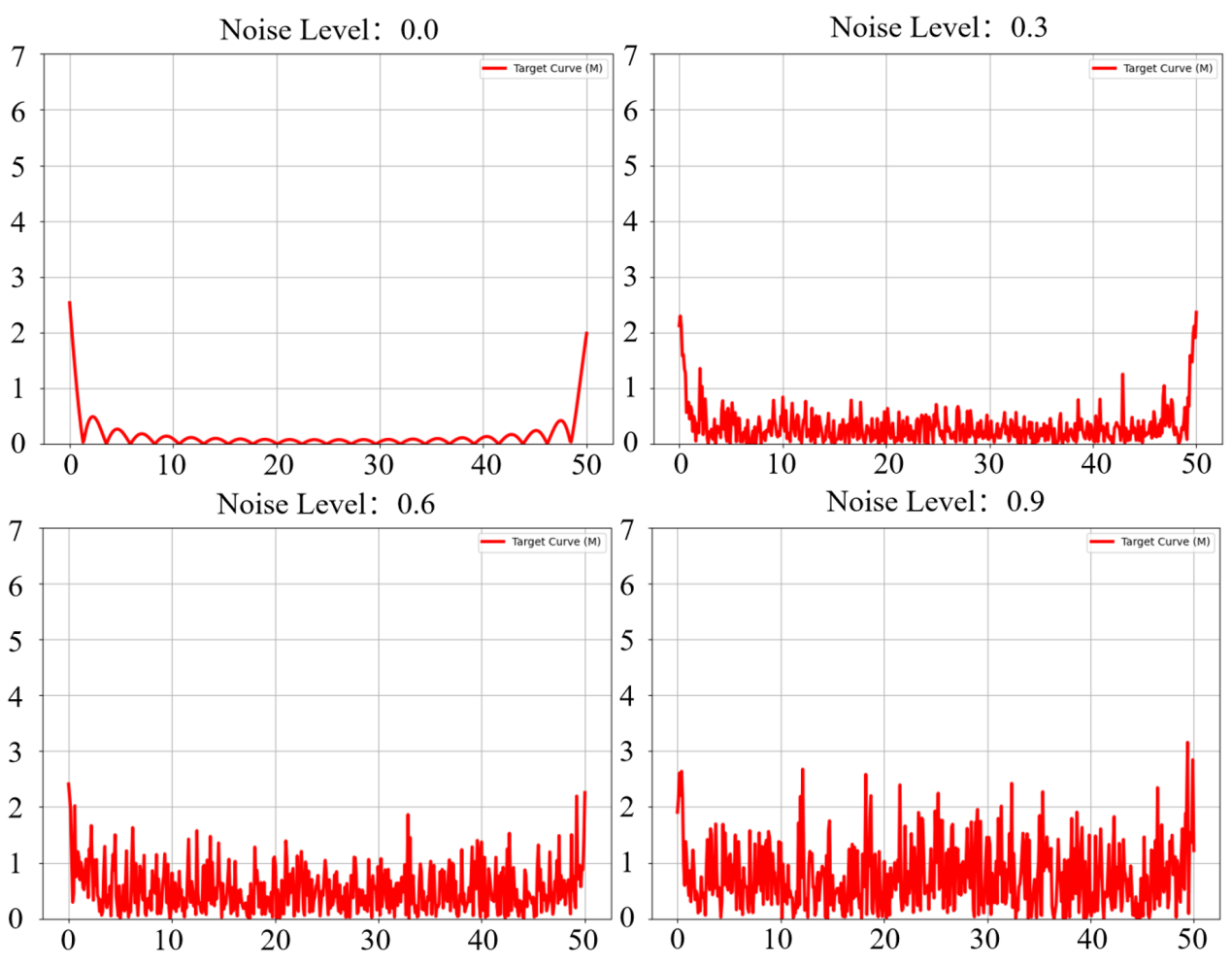}
        \caption{High-frequency information curve}
        \label{fig4.2}
    \end{subfigure}

    \caption{Low- and high-frequency 
    components of the target curves under varying levels of gaussian noise.}
    \label{fig4}
\end{figure}

\begin{figure}[!t]
\centerline{\includegraphics[width=\columnwidth]{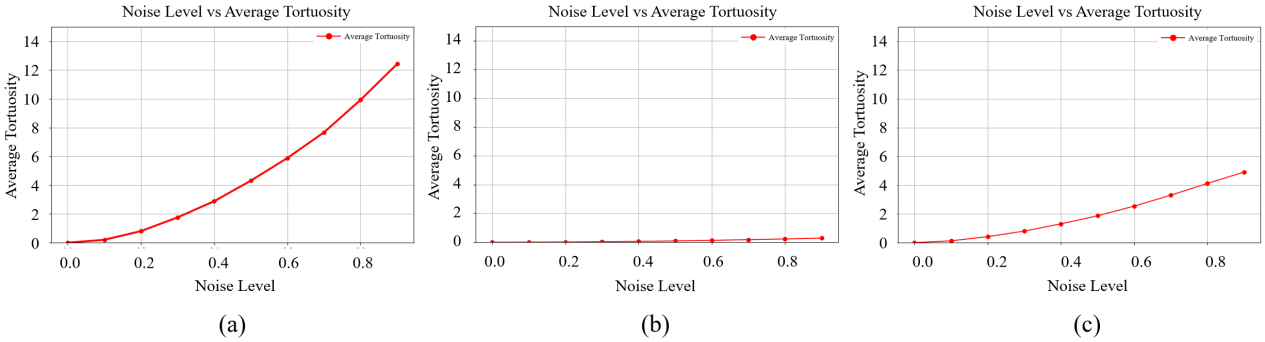}}
\caption{Tortuosity variation curves of the target curves and their low- 
and high-frequency components under different levels of gaussian noise.
(a) Original target curve 
(b) Low-frequency information curve
(c) High-frequency information curve}
\label{fig5}
\end{figure}

First, a sine curve was generated as the reference curve. Then, 
Gaussian random noise with a mean ($\mu$) of 0 and SDs ($\sigma$) ranging 
from 0.0 to 0.9 at 10 different levels was added to the curve. For each noise level, 
5000 random target curves were generated for experimental validation.
The average computation time for curvature evaluation per curve was approximately 0.0356 s.
Figure \ref{fig3} illustrates the selected 
standard curves (in blue) and corresponding target curves (in red) 
at four different noise levels. Figure \ref{fig4.1} shows the low-frequency 
components of the target curves at various noise levels, whereas Fig. \ref{fig4.2} 
presents the corresponding high-frequency components. Specifically, 
when the noise level is 0.0, the curves represent the low- 
or high-frequency characteristics of the standard curve. 
The average tortuosity of all target curves was calculated using the IEBF method, 
and the trend across different levels of Gaussian noise was plotted, as shown in Fig. \ref{fig5}a.
The tortuosity variations of the low- and 
high-frequency components of the target curves under different noise 
intensities are shown in Figs. \ref{fig5}b and \ref{fig5}c, respectively.

As shown in Fig. \ref{fig5}a, the tortuosity of the target curve 
increases monotonically with the noise level, indicating that the 
proposed method can effectively quantify the tortuosity of target 
data relative to the standard curve. When Gaussian noise is added 
at the highest level (Noise Level = 0.9), the tortuosity of the 
target curve reaches 12.43. Figures. \ref{fig5}b and \ref{fig5}c show that the 
tortuosity calculated from both low- and high-frequency 
components increases gradually with noise level. Because Gaussian noise 
has minimal impact on the low-frequency components, the maximum tortuosity 
calculated from the low-frequency information is only 0.53. This method 
not only effectively quantifies the influence of Gaussian noise on curve 
tortuosity, but also leverages Fourier transform to filter out high-frequency 
noise interference, thereby enhancing the robustness of structural analysis.

 \subsection{Experiment on Meibomian Gland Atrophy Uniformity}
 In this study, 105 subjects, comprising 35 patients 
 in the Demodex-negative group and 70 in the Demodex-positive group, 105 subjects were included, with ages 
 ranging from 26 to 45 years. The average computation time for the atrophy 
 uniformity index per image was approximately 0.6267 s. 
 The Mann-Whitney U test was used to analyze the 
 differences in the gland atrophy uniformity indices between the two groups. 
 The results indicated a statistically significant difference in uniformity 
 indices between the negative and positive groups ($p < 0.05$). The basic 
 demographic information, descriptive statistics, and $p$-values from 
 the Mann-Whitney U test are summarized in Table~\ref{table1}.

\begin{figure}[!t]
    \centering
    \begin{subfigure}[b]{0.48\columnwidth}
        \centering
        \includegraphics[width=\linewidth]{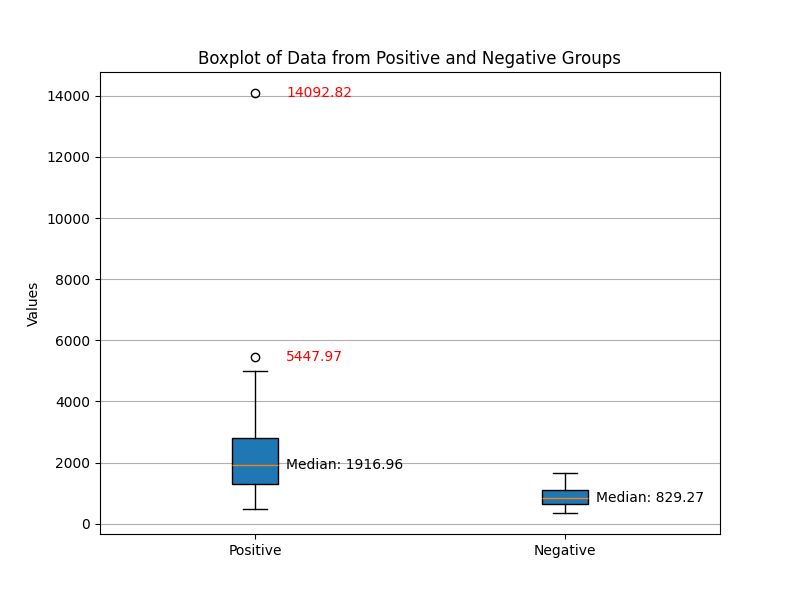}
        \caption{}
        \label{fig6.1}
    \end{subfigure}
    \hfill
    \begin{subfigure}[b]{0.48\columnwidth}
        \centering
        \includegraphics[width=\linewidth]{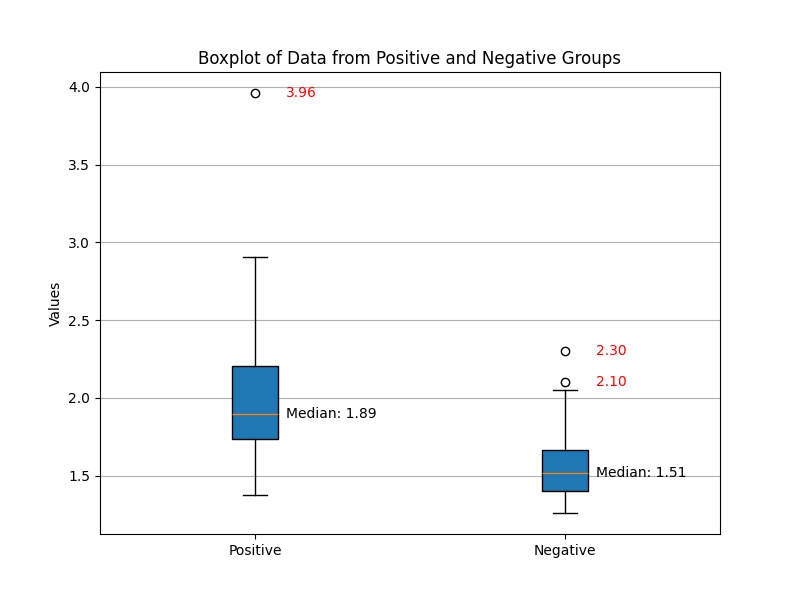}
        \caption{}
        \label{fig6.2}
    \end{subfigure}
    \hfill
    \begin{subfigure}[b]{0.48\columnwidth}
        \centering
        \includegraphics[width=\linewidth]{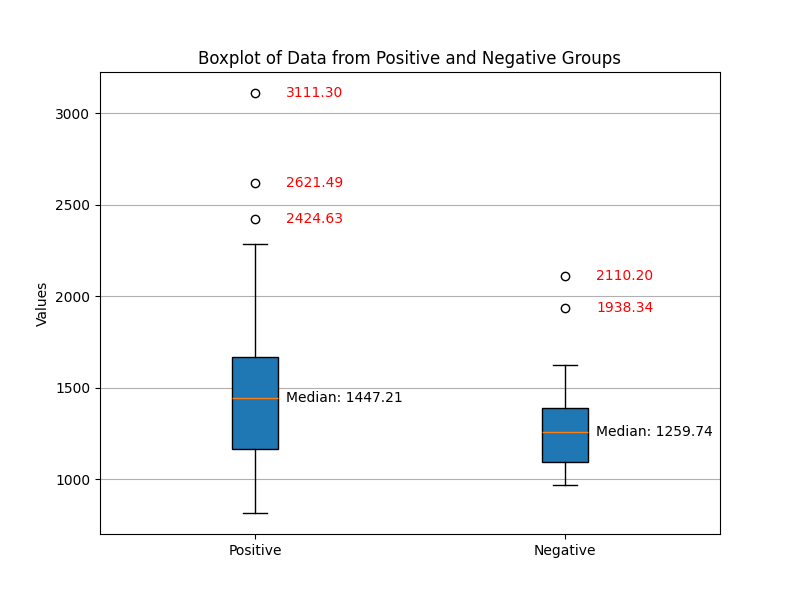}
        \caption{}
        \label{fig6.3}
    \end{subfigure}
    
    \caption{Box plots of tortuosity results for the normal control and abnormal groups.
    (a)IEBF (our method), (b) chord-to-arc ratio method, (c)total variation divided by gland density method.}
    \label{fig6}
\end{figure}

\begin{figure}[!t]
    \centering
    \begin{subfigure}[b]{0.48\columnwidth}
        \centering
        \includegraphics[width=\linewidth]{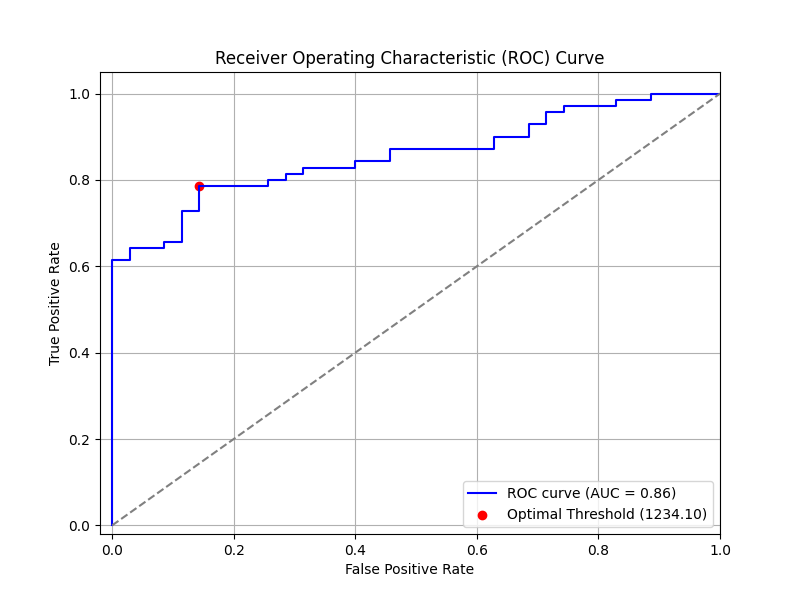}
        \caption{}
        \label{fig7.1}
    \end{subfigure}
    \hfill
    \begin{subfigure}[b]{0.48\columnwidth}
        \centering
        \includegraphics[width=\linewidth]{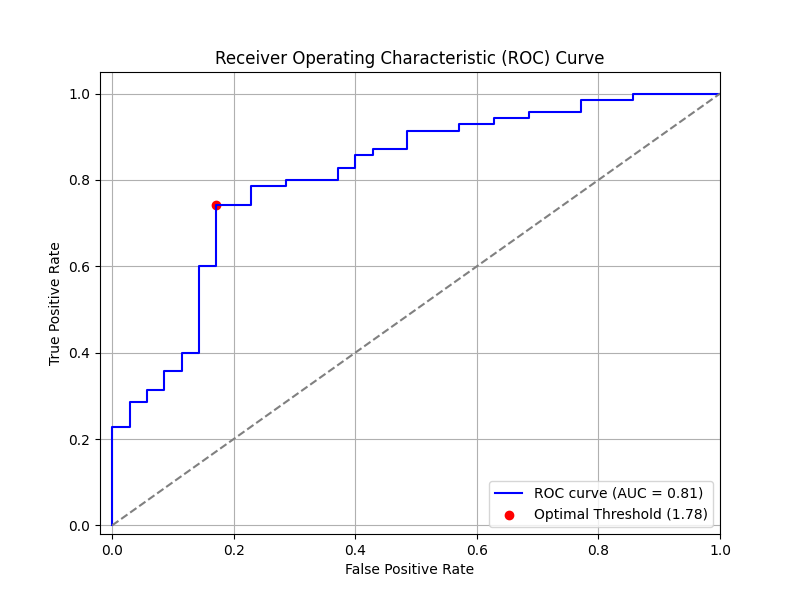}
        \caption{}
        \label{fig7.2}
    \end{subfigure}
    \hfill
    \begin{subfigure}[b]{0.48\columnwidth}
        \centering
        \includegraphics[width=\linewidth]{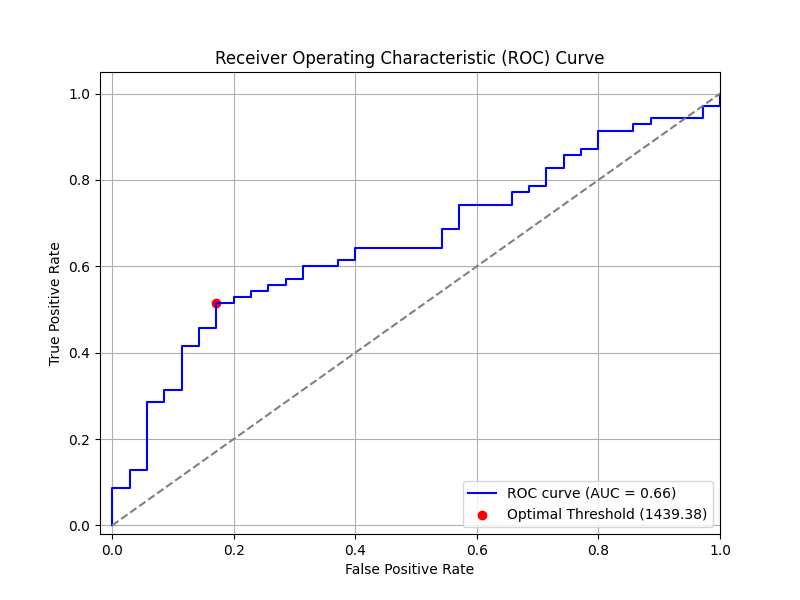}
        \caption{}
        \label{fig7.3}
    \end{subfigure}
    
    \caption{ROC curve of tortuosity distinguishing the normal control group from the abnormal group.
    (a)Our method, (b) chord-to-arc ratio method, and (c)total variation divided by gland density method.}
    \label{fig7}
\end{figure}

This study employed box plots and ROC curves to 
visualize the differences in key metrics between the negative and positive groups. 
As shown in Fig. \ref{fig6}, the box plots clearly illustrate the distributional differences 
between the two groups across the three methods. Furthermore, Fig. \ref{fig7} presents the 
ROC curves of the uniformity indices derived from each method. The corresponding AUC, sensitivity, specificity, and optimal classification thresholds 
are summarized in Table~\ref{table2}. 
The results demonstrate that IEBF achieved the best performance 
in distinguishing between the two groups, with an AUC, sensitivity, and specificity of 0.858, 0.786, 
and 0.857,respectively, significantly outperforming the other methods and exhibiting 
superior discriminative capability.

%%%%%%%%%%%%%%%%%%%%%%%%%%%%%%%%%%%%%%%%%%%%%%%%%%%%%%%%%%%%%%%%%%%%%%%%%%%%%%%%%%%%%%%%%%%%%
%%%%%%%%%%%%%%%%%%%%%%%%%%%%%%%%%%%%%%%%%%%%%%%%%%%%%%%%%%%%%%%%%%%%%%%%%%%%%%%%%%%%%%%%%%%%%
\begin{table}
    \caption{Comparison of Clinical and Quantitative Indicators Between \textit{Demodex} Positive and Negative Groups}
    \setlength{\tabcolsep}{3pt}
    \begin{tabular}{|p{80pt}|p{60pt}|p{60pt}|p{10pt}|}
        \hline
         & \textit{Demodex} Negative & \textit{Demodex} Positive& p \\
        \hline
        Sex & M=11(31.4\%) F=24(68.6\%)& M=19(27.1\%) F=51(72.9\%)  & 0.647 \\
        Age (Years (Mean ± Std)) & 35.37±9.30 & 36.64±8.20 & 0.289 \\
        IEBF (Our Method) (Median (IQR)) & 829.27 (645.51-1094.64) & 1916.96 (1298.12-2790.92)& <0.001 \\
        Chord-to-Arc Ratio (Median (IQR)) & 1.52(1.39-1.66) & 1.89(1.74-2.21) & <0.001 \\
        Total Variation (Median (IQR)) & 1259.74 (1095.39-1391.53) & 1447.21 (1165.76-1667.55) & 0.009 \\

        \hline
        \multicolumn{4}{p{250pt}}{M indicates the number and percentage of male participants, 
        and F indicates those of female participants. 'IEBF', 'Chord-to-Arc Ratio', 
        and 'Total Variation' represent the median and interquartile range (IQR) of 
        tortuosity values calculated using the corresponding methods. }
        
    \end{tabular}
    \label{table1}
\end{table}

%%两个表格

\begin{table}
    \caption{Performance Comparison of Different Methods Across Various Metrics}
    \setlength{\tabcolsep}{3pt}
    \begin{tabular}{|p{80pt}|p{55pt}|p{40pt}|p{40pt}|}
        \hline
         & AUC(95\%CI) & Sensitivity & Specificity \\
        \hline
        IEBF (Our Method) & 0.858(0.782-0.921) 
        & 0.786 & 0.857 \\
        Chord-to-Arc Ratio & 0.811(0.705-0.899) & 0.743 & 0.829 \\
        Total Variation & 0.656(0.548-0.760) & 514 & 0.829 \\

        \hline
    \end{tabular}
    \label{table2}
\end{table}
%%%%%%%%%%%%%%%%%%%%%%%%%%%%%%%%%%%%%%%%%%%%%%%%%%%%%%%%%%%%%%%%%%%%%%%%%%%%%%%%%%%%%%%%%%%%%
%%%%%%%%%%%%%%%%%%%%%%%%%%%%%%%%%%%%%%%%%%%%%%%%%%%%%%%%%%%%%%%%%%%%%%%%%%%%%%%%%%%%%%%%%%%%%

\section{Discussion}
This study proposes a novel tortuosity evaluation metric based on information entropy, 
which was validated and applied through numerical experiments and a clinical case study 
of \textit{Demodex} blepharitis. Currently, the diagnosis of Demodex blepharitis primarily relies 
on clinical signs such as lid margin hyperemia, scaling, cylindrical dandruff, eyelash 
abnormalities, and meibomian gland atrophy; however, these manifestations lack sufficient 
specificity\cite{yeu2025meibomian}. 
The current gold standard for diagnosing \textit{Demodex} infection includes microscopic 
examination of epilated eyelashes and confocal microscopy observation of the 
eyelash roots and eyelid margin. Particularly, confocal microscopy offers 
non-invasive, high-resolution imaging to detect \textit{Demodex} mites and associated 
tissue responses\cite{randon2015confocal, wang2019prospective}. 
However, this technique is operationally complex, 
requires specialized equipment and experienced clinicians, and entails lengthy 
examination times and high costs, limiting its widespread adoption in routine 
ophthalmic practice. Consequently, clinical diagnosis often depends on subjective 
assessments, highlighting the need for a more convenient, objective, and 
cost-effective quantitative method for structural alteration assessment.

Previous studies have suggested that uneven atrophy of the meibomian glands 
may represent a structural hallmark of Demodex infestation, which can be characterized 
by the curvature of the glandular boundary contour. This feature may help differentiate 
Demodex-related dysfunction from other types of meibomian gland abnormalities. However, 
no standardized quantitative criteria have been established to incorporate this 
characteristic into clinical diagnostic workflows, thereby limiting its potential 
application in practice\cite{liu2022uneven, yu2022uneven}.

To accurately quantify the structural complexity of curves, various tortuosity  
evaluation methods have been proposed in related studies across multiple disciplines, 
including ophthalmology, vascular surgery, and neurosurgery. 
The chord-to-arc ratio method\cite{lin2020novel, wang2017extracranial, 
zebic2023coronary, cinti2025differences}, widely 
favored for computational simplicity, measures overall tortuosity as the ratio 
of curve length to chord length; however, it fails to capture local detail and multi-scale 
structural variations of boundary curves. Additionally, it is sensitive to image 
resolution and noise, which impairs clinical applicability. The total 
variation method\cite{liu2022uneven} quantifies overall curve morphology by assessing 
gradient fluctuations but similarly exhibits noise sensitivity. Curvature-based 
methods\cite{mao2020automated, smedby2008two, hart1999measurement, sharafi2024automated,
turior2013quantification, patasius2005evaluation, kashyap2022accuracy, dasilva2022analysis, 
wang2017extracranial, badawi2022four} provide precise local bending characterization; however, they rely 
heavily on second-order derivatives, rendering them vulnerable to noise-induced 
instability and limiting their representation of global morphology. In anatomical 
structures exhibiting natural physiological tortuosity (e.g., meibomian glands, 
blood vessels, nerves), tortuosity metrics often generate excessive local extrema, 
complicating differentiation between physiological and pathological variations. 
Moreover, these methods generally reference ideal straight lines, which 
inadequately reflect the nonlinear structural features prevalent in biological 
tissues, often misclassifying normal tortuosity as pathological.

To address these challenges, we introduce an information entropy-based curve 
tortuosity quantification framework. This approach leverages Fourier transform-based 
frequency domain decomposition to separate low- and high-frequency components of 
the target curve, effectively suppressing high-frequency noise interference and 
enhancing robustness and stability in tortuosity estimation. Building on this, 
the framework employs information entropy coupled with probabilistic modeling 
to quantify the “disorder” in structural differences between the target and 
standard curves, thereby providing a comprehensive characterization of tortuosity. 
Unlike conventional approaches anchored to straight-line references, this 
framework enables flexible selection of physiologically relevant standard curves, 
aligning evaluations with actual anatomical morphology and improving discrimination 
between pathological and physiological tortuosity changes.

Information entropy, a well-established theoretical measure of system uncertainty 
and complexity, has been extensively applied in biomedical data analysis including 
image compression, texture recognition, signal processing, 
and electroencephalography\cite{srinivasan2007approximate, ramos2019computational, 
li2023comparison, kattumannil2022generalized}. Its inherent advantage in capturing 
structural complexity makes it a natural candidate for curve tortuosity assessment. 
Incorporating information entropy into tortuosity quantification not only overcomes 
the dependency on fixed geometric references inherent in traditional methods; however, it 
also confers enhanced generalizability, sensitivity, and interpretability. 
This offers a novel quantitative tool for analyzing structural variations 
in medical imaging, with promising clinical applications.

The proposed framework demonstrated strong performance in numerical 
simulation experiments. Furthermore, when applied to clinical cases of 
\textit{Demodex} blepharitis, it achieved promising discriminative results. 
Specifically, the diagnostic performance for \textit{Demodex} blepharitis infection 
yielded an AUC of 0.858, with a sensitivity of 0.786 and a specificity of 0.857, 
providing an accurate and objective quantitative indicator for evaluating 
the uneven atrophy of Meibomian glands and enhancing diagnostic efficiency. 
However, during the study, it was observed that the accuracy of Meibomian 
gland region segmentation had a significant impact on the discriminative 
power of the proposed index, with segmentation errors contributing to a 
reduction in overall classification performance.

However, the proposed evaluation framework still has several limitations. 
First, it relies heavily on the selection of a reference (standard) curve; 
inappropriate construction of the standard curve may affect both the accuracy 
and generalizability of the assessment. Second, while the Fourier transform 
is effective for separating frequency components, it has inherent limitations 
when applied to non-stationary signals or abrupt local variations, potentially 
leading to loss or blurring of critical local details, thereby impacting the 
fine recognition of complex boundaries. In addition, the computation of information 
entropy is relatively resource-intensive, which may hinder real-time application 
in large-scale clinical datasets. Lastly, further validation across diverse clinical 
settings and datasets is necessary to ensure the robustness and broad applicability 
of the proposed framework.

In the future, we aim to further improve the proposed framework by optimizing 
the construction of the standard curve. Specifically, we plan to enable the 
standard curve to predict subsequent trends based on the historical trajectory 
of the target curve. By quantifying the degree of disorder between the actual 
curve and the predicted trend, we will address the challenge of standard 
curve selection more effectively. In addition, to overcome the high computational 
complexity of entropy calculation, we will explore algorithmic optimizations and 
acceleration strategies to enhance processing efficiency, thereby supporting rapid 
analysis and real-time application in large-scale clinical datasets. Finally, we 
will the clinical applicability of this framework by integrating 
multimodal medical imaging data and conducting disease-specific investigations 
across diverse clinical domains. Through these advancements, the proposed tortuosity 
evaluation framework holds promise for widespread use in 
ophthalmology\cite{chiang2021international, sharafi2024automated, 
patasius2005evaluation, badawi2022four}, 
neurosurgery\cite{wang2017extracranial}, 
vascular surgery\cite{zebic2023coronary, huang2024computer}, 
and other clinical disciplines, providing a more 
reliable quantitative tool for early diagnosis and personalized treatment.

\section{Conclusion}
This study proposes a novel curve tortuosity quantification framework that integrates 
probabilistic modeling and information entropy theory, incorporating frequency-domain 
transformation of curve data to achieve an objective quantification of structural complexity. 
Unlike traditional tortuosity or chord-to-arc ratio methods, this approach does not 
rely on ideal straight lines as references; instead, it quantifies the “disorder” 
in structural differences by comparing target curves with standard reference curves, 
thereby providing a more robust and adaptable tortuosity metric. The framework demonstrates 
strong stability in numerical simulations and has been successfully applied to 
quantitatively analyze uneven meibomian gland atrophy in patients with \textit{Demodex} blepharitis, 
exhibiting excellent discriminative capability. This work extends the application of 
information entropy and probabilistic models in medical curve analysis, offering novel 
insights and methodologies for quantifying structural changes in clinical imaging, 
with promising potential for broader adoption.

\section*{References}

\bibliographystyle{IEEEtran}       % 或其他样式
\bibliography{references}

\begin{thebibliography}{10}
\providecommand{\url}[1]{#1}
\csname url@samestyle\endcsname
\providecommand{\newblock}{\relax}
\providecommand{\bibinfo}[2]{#2}
\providecommand{\BIBentrySTDinterwordspacing}{\spaceskip=0pt\relax}
\providecommand{\BIBentryALTinterwordstretchfactor}{4}
\providecommand{\BIBentryALTinterwordspacing}{\spaceskip=\fontdimen2\font plus
\BIBentryALTinterwordstretchfactor\fontdimen3\font minus
  \fontdimen4\font\relax}
\providecommand{\BIBforeignlanguage}[2]{{%
\expandafter\ifx\csname l@#1\endcsname\relax
\typeout{** WARNING: IEEEtran.bst: No hyphenation pattern has been}%
\typeout{** loaded for the language `#1'. Using the pattern for}%
\typeout{** the default language instead.}%
\else
\language=\csname l@#1\endcsname
\fi
#2}}
\providecommand{\BIBdecl}{\relax}
\BIBdecl

\bibitem{chiang2021international}
M.~F. Chiang, G.~E. Quinn, A.~R. Fielder, S.~R. Ostmo, R.~V. Paul~Chan,
  A.~Berrocal \emph{et~al.}, ``International classification of retinopathy of
  prematurity, third edition,'' \emph{Ophthalmology}, vol. 128, no.~10, pp.
  e51--e68, 2021.

\bibitem{you2022application}
A.~You, J.~K. Kim, I.~H. Ryu, and T.~K. Yoo, ``Application of generative
  adversarial networks (gan) for ophthalmology image domains: a survey,''
  \emph{Eye and Vision}, vol.~9, no.~1, p.~6, 2022.

\bibitem{williams2020artificial}
B.~M. Williams, D.~Borroni, R.~Liu, Y.~Zhao, J.~Zhang, J.~Lim \emph{et~al.},
  ``An artificial intelligence-based deep learning algorithm for the diagnosis
  of diabetic neuropathy using corneal confocal microscopy: a development and
  validation study,'' \emph{Diabetologia}, vol.~63, no.~2, pp. 419--430, 2020.

\bibitem{wei2020deep}
S.~Wei, F.~Shi, Y.~Wang, Y.~Chou, and X.~Li, ``A deep learning model for
  automated sub-basal corneal nerve segmentation and evaluation using in vivo
  confocal microscopy,'' \emph{Translational Vision Science \& Technology},
  vol.~9, no.~2, p.~32, 2020.

\bibitem{cruzat2017vivo}
A.~Cruzat, Y.~Qazi, and P.~Hamrah, ``In vivo confocal microscopy of corneal
  nerves in health and disease,'' \emph{Ocular Surface}, vol.~15, no.~1, pp.
  15--47, 2017.

\bibitem{gao2005in}
Y.~Y. Gao, M.~A. Di~Pascuale, W.~Li, A.~Baradaran-Rafii, A.~Elizondo, C.~L. Kuo
  \emph{et~al.}, ``In vitro and in vivo killing of ocular demodex by tea tree
  oil,'' \emph{British Journal of Ophthalmology}, vol.~89, no.~11, pp.
  1468--1473, 2005.

\bibitem{kheirkhah2007corneal}
A.~Kheirkhah, V.~Casas, W.~Li, V.~K. Raju, and S.~C. Tseng, ``Corneal
  manifestations of ocular demodex infestation,'' \emph{American Journal of
  Ophthalmology}, vol. 143, no.~5, pp. 743--749, 2007.

\bibitem{lee2010relationship}
S.~H. Lee, Y.~S. Chun, J.~H. Kim, E.~S. Kim, and J.~C. Kim, ``The relationship
  between demodex and ocular discomfort,'' \emph{Investigative Ophthalmology \&
  Visual Science}, vol.~51, no.~6, pp. 2906--2911, 2010.

\bibitem{cheng2019correlation}
S.~Cheng, M.~Zhang, H.~Chen, W.~Fan, and Y.~Huang, ``The correlation between
  the microstructure of meibomian glands and ocular demodex infestation: A
  retrospective case-control study in a chinese population,'' \emph{Medicine
  (Baltimore)}, vol.~98, no.~19, 2019.

\bibitem{randon2015confocal}
M.~Randon, H.~Liang, M.~El~Hamdaoui, R.~Tahiri, L.~Batellier, A.~Denoyer
  \emph{et~al.}, ``In vivo confocal microscopy as a novel and reliable tool for
  the diagnosis of demodex eyelid infestation,'' \emph{British Journal of
  Ophthalmology}, vol.~99, no.~3, pp. 336--341, 2015.

\bibitem{alver2017clinical}
O.~Alver, S.~A. Kivanc, B.~Akova~Budak, N.~U. Tuzemen, B.~Ener, and A.~T.
  Ozmen, ``A clinical scoring system for diagnosis of ocular demodicosis,''
  \emph{Medical Science Monitor}, vol.~23, pp. 5862--5869, 2017.

\bibitem{zhang2018association}
X.~B. Zhang, Y.~H. Ding, and W.~He, ``The association between demodex
  infestation and ocular surface manifestations in meibomian gland
  dysfunction,'' \emph{International Journal of Ophthalmology}, vol.~11, no.~4,
  pp. 589--592, 2018.

\bibitem{yan2020association}
Y.~Yan, Q.~Yao, Y.~Lu, C.~Shao, H.~Sun, Y.~Li \emph{et~al.}, ``Association
  between demodex infestation and ocular surface microbiota in patients with
  demodex blepharitis,'' \emph{Frontiers in Medicine}, vol.~7, p. 592759, 2020.

\bibitem{liang2014high}
L.~Liang, X.~Ding, and S.~C. Tseng, ``High prevalence of demodex brevis
  infestation in chalazia,'' \emph{American Journal of Ophthalmology}, vol.
  157, no.~2, pp. 342--348.e1, 2014.

\bibitem{hao2022demodex}
Y.~Hao, X.~Zhang, J.~Bao, L.~Tian, and Y.~Jie, ``Demodex folliculorum
  infestation in meibomian gland dysfunction related dry eye patients,''
  \emph{Frontiers in Medicine}, vol.~9, p. 833778, 2022.

\bibitem{saha2022automated}
R.~K. Saha, A.~M.~M. Chowdhury, K.~S. Na, G.~D. Hwang, Y.~Eom, J.~Kim
  \emph{et~al.}, ``Automated quantification of meibomian gland dropout in
  infrared meibography using deep learning,'' \emph{Ocular Surface}, vol.~26,
  pp. 283--294, 2022.

\bibitem{li2023unsupervised}
S.~Li, Y.~Wang, C.~Yu, Q.~Li, P.~Chang, D.~Wang \emph{et~al.}, ``Unsupervised
  learning based on meibography enables subtyping of dry eye disease and
  reveals ocular surface features,'' \emph{Investigative Ophthalmology \&
  Visual Science}, vol.~64, no.~13, p.~43, 2023.

\bibitem{liu2022uneven}
X.~Liu, Y.~Fu, D.~Wang, S.~Huang, C.~He, X.~Yu \emph{et~al.}, ``Uneven index: A
  digital biomarker to prompt demodex blepharitis based on deep learning,''
  \emph{Frontiers in Physiology}, vol.~13, p. 934821, 2022.

\bibitem{lin2020novel}
X.~Lin, Y.~Fu, L.~Li, C.~Chen, X.~Chen, Y.~Mao \emph{et~al.}, ``A novel
  quantitative index of meibomian gland dysfunction, the meibomian gland
  tortuosity,'' \emph{Translational Vision Science \& Technology}, vol.~9,
  no.~9, p.~34, 2020.

\bibitem{mao2020automated}
J.~Mao, Y.~Luo, L.~Liu, J.~Lao, Y.~Shao, M.~Zhang \emph{et~al.}, ``Automated
  diagnosis and quantitative analysis of plus disease in retinopathy of
  prematurity based on deep convolutional neural networks,'' \emph{Acta
  Ophthalmologica}, vol.~98, no.~3, pp. e339--e345, 2020.

\bibitem{smedby2008two}
{\"O}.~Smedby, N.~H{\"o}gman, S.~Nilsson, U.~Erikson, A.~G. Olsson, and
  G.~Walldius, ``Two-dimensional tortuosity of the superficial femoral artery
  in early atherosclerosis,'' \emph{Journal of Vascular Research}, vol.~30,
  no.~4, pp. 181--191, 2008.

\bibitem{hart1999measurement}
W.~E. Hart, M.~Goldbaum, B.~Côté, P.~Kube, and M.~R. Nelson, ``Measurement
  and classification of retinal vascular tortuosity,'' \emph{International
  Journal of Medical Informatics}, vol.~53, no.~2, pp. 239--252, 1999.

\bibitem{sharafi2024automated}
S.~M. Sharafi, N.~Ebrahimiadib, R.~Roohipourmoallai, A.~D. Farahani, M.~I.
  Fooladi, and E.~Khalili~Pour, ``Automated diagnosis of plus disease in
  retinopathy of prematurity using quantification of vessels characteristics,''
  \emph{Scientific Reports}, vol.~14, no.~1, p. 6375, 2024.

\bibitem{turior2013quantification}
R.~Turior, D.~Onkaew, B.~Uyyanonvara, and P.~Chutinantvarodom, ``Quantification
  and classification of retinal vessel tortuosity,'' \emph{ScienceAsia},
  vol.~39, no.~3, 2013.

\bibitem{patasius2005evaluation}
M.~Patasius, V.~Marozas, A.~Lukosevicius, and D.~Jegelevicius, ``Evaluation of
  tortuosity of eye blood vessels using the integral of square of derivative of
  curvature,'' in \emph{IFMBE Proceedings of the 3rd European Medical and
  Biological Engineering Conference (EMBEC05)}, vol.~11, 2005.

\bibitem{kashyap2022accuracy}
V.~Kashyap, R.~Gharleghi, D.~D. Li, L.~McGrath-Cadell, R.~M. Graham, C.~Ellis
  \emph{et~al.}, ``Accuracy of vascular tortuosity measures using computational
  modelling,'' \emph{Scientific Reports}, vol.~12, no.~1, p. 865, 2022.

\bibitem{dasilva2022analysis}
M.~V. da~Silva, J.~Ouellette, B.~Lacoste, and C.~H. Comin, ``An analysis of the
  influence of transfer learning when measuring the tortuosity of blood
  vessels,'' \emph{Computer Methods and Programs in Biomedicine}, vol. 225, p.
  107021, 2022.

\bibitem{wang2024improved}
Y.~Wang, X.~Zheng, C.~He, J.~Zhang, and S.~Huang, ``Improved segmentation of
  infant retinal images and quantitative vascular analysis,'' \emph{IEEE
  Access}, vol.~12, pp. 181\,846--181\,857, 2024.

\bibitem{wang2017extracranial}
H.~F. Wang, D.~M. Wang, J.~J. Wang, L.~J. Wang, J.~Lu, P.~Qi \emph{et~al.},
  ``Extracranial internal carotid artery tortuosity and body mass index,''
  \emph{Frontiers in Neurology}, vol.~8, p. 508, 2017.

\bibitem{badawi2022four}
S.~A.~Q. Badawi, M.~Takruri, Y.~Albadawi, M.~A.~K. Khattak, A.~K. Nileshwar,
  and E.~Mosalam, ``Four severity levels for grading the tortuosity of a
  retinal fundus image,'' \emph{Journal of Imaging}, vol.~8, no.~10, 2022.

\bibitem{zebic2023coronary}
P.~Zebic~Mihic, J.~Arambasic, D.~Mlinarevic, S.~Saric, M.~Labor, I.~Bosnjak
  \emph{et~al.}, ``Coronary tortuosity index vs. angle measurement method for
  the quantification of the tortuosity of coronary arteries in non-obstructive
  coronary disease,'' \emph{Diagnostics (Basel)}, vol.~14, no.~1, 2023.

\bibitem{krivosic2025assessment}
V.~Krivosic, P.~Goupillou, F.~Buffon-Porcher, H.~Morel, S.~Guey, R.~Tadayoni
  \emph{et~al.}, ``Assessment of retinal arteriolar tortuosity in patients with
  col4a1 or col4a2 mutations,'' \emph{Retina}, vol.~45, no.~2, pp. 296--302,
  2025.

\bibitem{huang2024computer}
Y.-P. Huang, S.~Vadloori, E.~Y.-C. Kang, Y.~Fukushima, R.~Takahashi, and W.-C.
  Wu, ``Computer-aided detection of retinopathy of prematurity severity
  assessment via vessel tortuosity measurement in preterm infants’ fundus
  images,'' \emph{Eye}, vol.~38, no.~17, pp. 3309--3317, 2024.

\bibitem{zhang2022meibomian}
Z.~H. Zhang, X.~Yu, Y.~Fu, X.~Chen, W.~Yang, and Q.~Dai, ``Meibomian gland
  density: An effective evaluation index of meibomian gland dysfunction based
  on deep learning and transfer learning,'' \emph{Journal of Clinical
  Medicine}, vol.~11, no.~9, p. 2396, 2022.

\bibitem{yeu2025meibomian}
E.~Yeu and C.~Koetting, ``Meibomian gland structure and function in patients
  with demodex blepharitis,'' \emph{Journal of Cataract and Refractive
  Surgery}, vol.~51, no.~5, pp. 359--365, 2025.

\bibitem{wang2019prospective}
Y.~J. Wang, M.~Ke, and X.~M. Chen, ``Prospective study of the diagnostic
  accuracy of the in vivo laser scanning confocal microscopy for ocular
  demodicosis,'' \emph{American Journal of Ophthalmology}, vol. 203, pp.
  46--52, 2019.

\bibitem{yu2022uneven}
X.~Yu, Y.~Fu, H.~Lian, D.~Wang, Z.~Zhang, and Q.~Dai, ``Uneven meibomian gland
  dropout in patients with meibomian gland dysfunction and demodex
  infestation,'' \emph{Journal of Clinical Medicine}, vol.~11, no.~17, 2022.

\bibitem{cinti2025differences}
N.~Cinti, P.~J. McKeegan, P.~J. Bazira, A.~Smith, P.~Maliakal, M.~Danciut
  \emph{et~al.}, ``Differences in internal carotid artery tortuosity in
  ruptured and unruptured anterior circulation aneurysms. a matched
  case-control study,'' \emph{Neurochirurgie}, vol.~71, no.~1, p. 101613, 2025.

\bibitem{srinivasan2007approximate}
V.~Srinivasan, C.~Eswaran, and N.~Sriraam, ``Approximate entropy-based
  epileptic eeg detection using artificial neural networks,'' \emph{IEEE
  Transactions on Information Technology in Biomedicine}, vol.~11, no.~3, pp.
  288--295, 2007.

\bibitem{ramos2019computational}
L.~Ramos, J.~Novo, J.~Rouco, S.~Romeo, M.~D. Alvarez, and M.~Ortega,
  ``Computational assessment of the retinal vascular tortuosity integrating
  domain-related information,'' \emph{Scientific Reports}, vol.~9, no.~1, p.
  19940, 2019.

\bibitem{li2023comparison}
C.-B. Li and Y.-L. Ye, ``A comparison of topological entropies for
  nonautonomous dynamical systems,'' \emph{Journal of Mathematical Analysis and
  Applications}, vol. 517, no.~2, 2023.

\bibitem{kattumannil2022generalized}
S.~K. Kattumannil, E.~P. Sreedevi, and N.~Balakrishnan, ``A generalized measure
  of cumulative residual entropy,'' \emph{Entropy}, vol.~24, no.~4, 2022.

\end{thebibliography}

\end{document}